\documentclass[cameraready]{Interspeech}

\title{From Black-Box to Clinical Insight: A Multi-Stage Explainable Framework for Speech-Based Cognitive Impairment Detection}

\author[affiliation={1}, orcid=0009-0005-0160-6113]{Yasaman}{Haghbin}
\author[affiliation={1}, orcid=0009-0009-1937-3396]{Sina}{Rashidi}
\author[affiliation={1}, orcid=0009-0007-0668-5169]{Ali}{Zolnour}
\author[affiliation={1}, orcid=0009-0004-5650-6374]{Fatemeh}{Taherinezhad}
\author[affiliation={1}, orcid=0009-0007-7551-3993]{Ali}{Fartoot}
\author[affiliation={1}, orcid=0009-0008-4284-9869]{Hossein}{Azadmaleki}
\author[affiliation={2}, orcid=0000-0003-0648-6702]{James M}{Noble}
\author[affiliation={3}, orcid=0009-0003-9769-5979]{Maryam}{Dadkhah}
\author[affiliation={2}, orcid=0000-0003-4484-2990, correspondingauthor]{Maryam}{Zolnoori}

\address{
    $^1$ Independent Researcher \\
    $^2$ Columbia University, United States \\
    $^3$ Chalmers University of Technology, Sweden
}

\email{hbn.yasaman@gmail.com, sinarashidi46@gmail.com, zonour97@gmail.com,	ftaherin00@gmail.com, fartoot.ali.80@gmail.com, hosein.azadmaleki@gmail.com, jn2054@cumc.columbia.edu, tirani@chalmers.se, mz2825@cumc.columbia.edu}


\usepackage{comment}
\usepackage{subcaption}

\begin{document}

\maketitle

\begin{abstract}
Speech-based cognitive impairment detection offers a noninvasive, accessible alternative to costly biomarker assays, yet transformer-based models remain clinically uninterpretable. We propose a multi-stage explainability framework that translates black-box transformer predictions into clinically grounded narratives by integrating SHapley Additive exPlanations (SHAP)-based token attribution, theory-informed linguistic features, and a four-stage LLM reasoning pipeline using LLaMA-3.1-70B-Instruct. Built on the SpeechCARE-Adaptive Gating Network multimodal screening model (F1 = 72.11\% on the NIA PREPARE benchmark), the framework maps model outputs to four cognitive-linguistic dimensions, including lexical richness, syntactic complexity, and semantic coherence. Physician evaluation on 70 stratified English samples demonstrated strong alignment with patient-level cognitive profiles, and a System Usability Scale score of 82/100 indicated high potential for clinical workflow integration.
\end{abstract}

\section{Introduction}
Cognitive impairment, including mild cognitive impairment (MCI) and Alzheimer’s disease (AD), poses an urgent public health challenge, with projected U.S. prevalence reaching 11–16 million cases by 2050 \cite{petersen2016mild}. Speech offers a uniquely accessible, noninvasive biomarker for early cognitive decline \cite{meilan2020changes}. Acoustic and linguistic cues reflect impairments in phonetic motor planning, language organization, executive functioning, and semantic memory \cite{petersen2018practice}.

Transformer-based speech and language models have significantly improved cognitive impairment detection performance \cite{zolnoori2026detecting, rashidi2025speechcura, 11462646, haghbin2026voice}, yet their black-box nature remains a major barrier to clinical translation. Clinicians require explanations that link model outputs to clinically meaningful language and speech patterns for individual patients. This need has driven growing interest in Explainable Artificial Intelligence (XAI).

Despite this importance, XAI in speech-based cognitive impairment detection remains limited. For instance, Iqbal et al. \cite{iqbal2024explainable} trained a Random Forest classifier to the ADReSS Benchmark dataset \cite{luz2021alzheimer} using Part-of-Speech distributions and lexical diversity, with SHapley Additive exPlanations (SHAP) \cite{ekanayake2022novel} used to quantify cohort-level linguistic feature importance. 

More recent work has applied XAI to transformer models. Ilias et al. \cite{ilias2022explainable} used LIME \cite{garreau2020explaining} to interpret a BERT model trained on ADReSS transcripts, characterizing linguistic differences between AD and non-AD speech, while Li et al. \cite{li2024useful} applied SHAP to a fine-tuned BERT model on ADReSSo \cite{luz21_interspeech} to rank influential tokens. Rezaii et al. \cite{rezaii2025voiceprints} further applied SHAP to an XGBoost model and LIME to a fine-tuned RoBERTa-base model, indicating model reliance on vague expressions (e.g., “I don’t remember”) as markers of cognitive impairment.

Despite these advances, existing approaches emphasize token-level attributions or handcrafted features and provide limited linkage to cognitive-linguistic mechanisms that clinicians use for assessment. SHAP highlights influential tokens through numerical importance scores but does not explain why these linguistic patterns reflect underlying cognitive processes, and such technical outputs limit clinical usability for clinicians who require plain-language justifications.

To address these limitations, we propose a multi-stage explainability framework that systematically connects transformer predictions to clinically interpretable linguistic narratives. The framework integrates three complementary components: (1) SHAP adapted for transformer architectures with hierarchical aggregation of subword attributions to word-level explanations; (2) complementary linguistic features capturing clinically relevant dimensions such as lexical richness and syntactic complexity; and (3) sequential Large Language Model (LLM) orchestration, a four-stage reasoning pipeline using LLaMA-3.1-70B-Instruct \cite{grattafiori2024llama} to translate model attributions and linguistic evidence into structured, plain-language clinical explanations.

This framework extends the SpeechCARE Adaptive Gating Network (AGF) \cite{azadmaleki2025speechcare}, a multimodal screening model that integrates acoustic and linguistic transformer representations via a Mixture-of-Experts–inspired fusion architecture. Both the screening model and the explainability framework were recognized in the NIA PREPARE Challenge (April 2025) \cite{zolnoori2025national}. All codes are available in the \href{https://github.com/SpeechCARE/Cognitive-Speech-Explainer-Interspeech2026}{GitHub}.

\section{Method}

\subsection{Data: PREPARE challenge dataset}
We used the National Institute on Aging (NIA) PREPARE benchmark dataset \cite{azadmaleki2025speechcare} comprising speech recordings from 2,058 participants (1,646 training, 412 testing) across three languages (English, Spanish, Mandarin). The cohort included 1,140 cognitively healthy controls, 268 with MCI, and 650 with AD. Recordings were collected from multiple corpora, truncated to a maximum of 30 seconds (mean = 27 seconds). A validation set was created from the training data by randomly selecting 20\% (329 participants) using stratified splits across diagnostic groups to ensure balanced representation.

\subsection{Preprocessing components}
Data preprocessing consisted of three steps:
\textit{(1) Age Stratification:} Age was discretized into three categories reflecting cognitive aging stages: mid-life (46–65 years), older adults (66–80 years), and elderly (80+ years). This encoding was selected based on preliminary experiments showing improved performance over continuous age representation in downstream classification tasks.
\textit{(2) Amplitude Normalization:} Peak normalization standardized signal amplitude across recordings. Each waveform was rescaled so the maximum absolute amplitude reached 0.95. 
\textit{(3) Automatic Speech Recognition:} Whisper-Large \cite{radford2023robust}, a multilingual transformer-based ASR model, generated word-level transcripts with strong open-source performance across languages and acoustic conditions. To ensure transcript quality, all incomplete or erroneous transcripts were manually reviewed and corrected.

\subsection{Model architecture}

\begin{figure}
    \centering
    \includegraphics[width=0.95\linewidth]{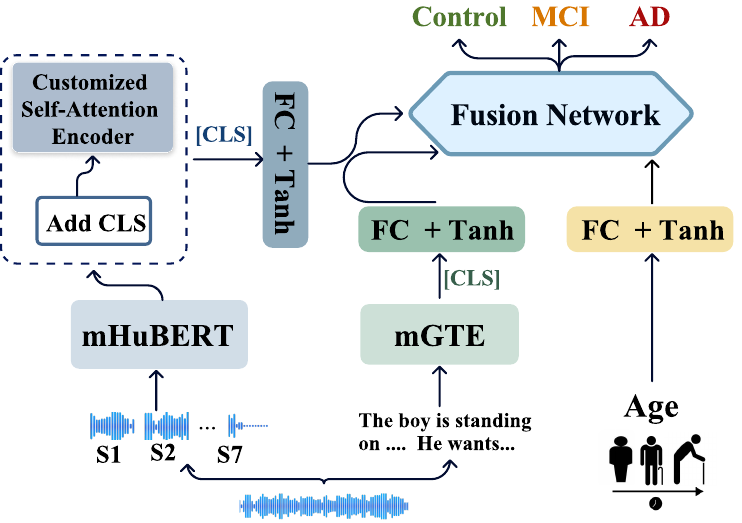}
    \caption{Architecture of the screening model. The model employed a gating mechanism to dynamically weight multimodal representations extracted from mGTE linguistic, mHuBERT acoustic encoders, and demographic age.}
    \label{fig:speechCARE}
\end{figure}

As a screening model, we employed SpeechCARE Adaptive Gating Fusion (SpeechCARE-AGF), a multimodal framework that dynamically assigned weights to different modalities. As depicted in Figure \ref{fig:speechCARE}, the system comprised two main components: a feature extraction network and a fusion network. Details of this model is available here \cite{azadmaleki2025speechcare}. 

\subsubsection{Feature network}
The feature network generated representations from pre-trained transformers specialized for language and speech. The linguistic encoder used mGTE \cite{zhang2024mgte}. The final-layer \texttt{[CLS]} token embedding provided the linguistic representation. The acoustic encoder processed audio through mHuBERT \cite{boito2024mhubert}. 

To accommodate 30-second recordings beyond the effective context of the acoustic encoder ($\sim$5\,s) while preserving local prosodic cues and modeling long-range temporal structure, audios were segmented into overlapping 5-second windows, encoded each segment with mHuBERT, and a trainable \texttt{[CLS]} token was added. The augmented sequence was processed through a customized self-attention encoder (two layers, four attention heads) to yield the acoustic representation. 

Age was encoded categorically into three groups (mid-life, older adults, elderly) as the demographic representation.

\subsubsection{Fusion network}
Inspired by Mixture-of-Experts \cite{lo2025closer} frameworks, the AGF network dynamically weighted modalities by attending to the most discriminative features. The hidden representations were concatenated and processed by a gating network that assigned dynamic weights, and each hidden representation was projected through a fully connected layer into a modality-specific output vector.  The weighted sum of these vectors formed the fused logits, and a SoftMax layer generated the final prediction. This architecture won the Special Recognition Prize in Phase 2 of the PREPARE Challenge.

\subsection{Model training and hyperparameter tuning}
To train the SpeechCARE model, both the mHuBERT and mGTE encoders were fine-tuned concurrently within a unified architecture. The model was trained for 15 epochs, with validation performance evaluated after each epoch and the best checkpoint (highest validation F1-score) selected for testing. We used learning rates of $10^{-6}$ for mGTE and $10^{-5}$ for the remaining components, and a batch size of 4. Fully connected layers used 128 neurons with Tanh activation, while the gating network employed 384 neurons (three modalities × 128 neurons each).

\subsection{Clinically-grounded explainability framework}
\begin{figure*}
    \centering
    \includegraphics[width=0.95\linewidth]{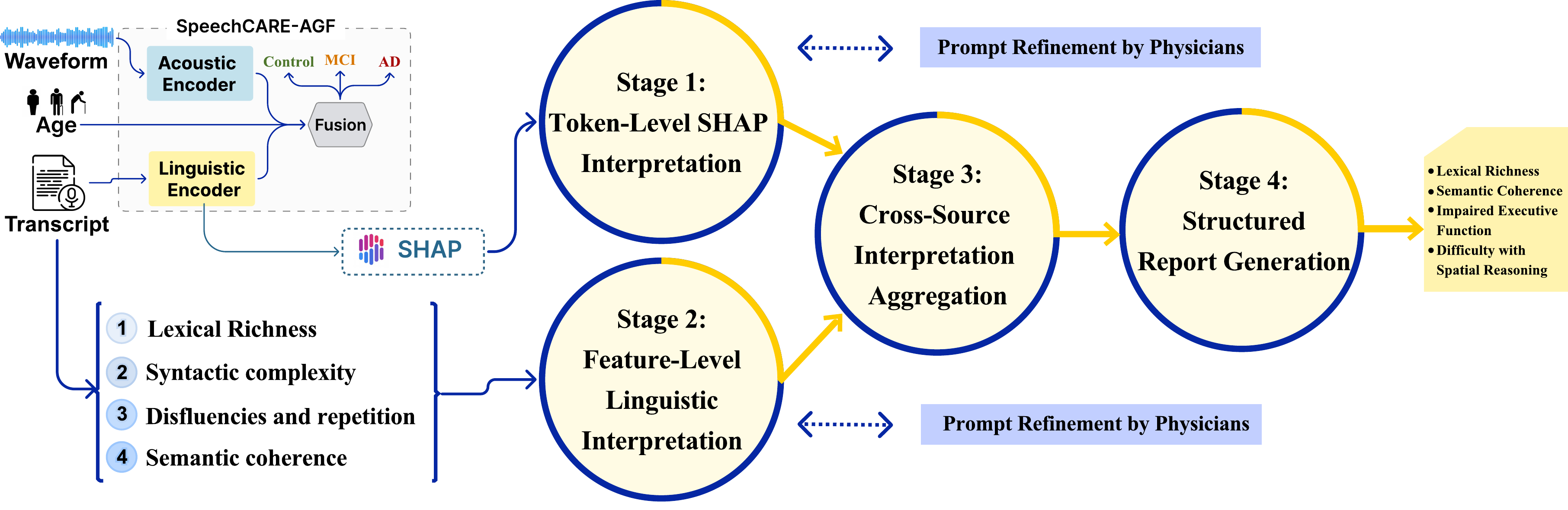}
    \caption{Multi-stage explainability pipeline for clinically grounded cognitive impairment interpretation. SHAP attributions are applied to the SpeechCARE-AGF screening model to extract token-level contributions. These attributions, alongside theory-informed linguistic features, are passed through four sequential LLM reasoning stages}
    \label{fig:abstract}
\end{figure*}

We propose a novel multi-stage explainability framework that link model attributions to clinically interpretable linguistic evidence. As shown in Figure \ref{fig:abstract}, our approach combines SHAP-based token attribution, theory-informed linguistic features, and LLM-powered reasoning to generate transparent, domain-relevant explanations for cognitive impairment detection.

\subsubsection{SHAP adaptation for transformer-based models}
Transformer-based models tokenize input text into subword units, complicating the application of perturbation-based explanation methods like SHAP \cite{li2024useful}. To address this, we implemented a model wrapper that encodes input text using the tokenizer, computes transformer embeddings, and outputs class probabilities, enabling SHAP to estimate the contribution of individual tokens to model predictions. To account for subword tokenization, we applied hierarchical SHAP value aggregation to map subword-level attributions back to interpretable word-level representations.

\subsubsection{Theory-informed linguistic feature extraction}
While SHAP values provide local token-level explanations, they lack explicit grounding in linguistically grounded interpretability. To address this limitation, we extracted handcrafted features across four clinically grounded domains \cite{zolnoori2023adscreen, zolnour2025llmcare}: (1) Lexical richness (e.g. type–token ratios, Brunet's and Honore's indices); (2) Syntactic complexity (e.g. average clause length, part-of-speech diversity); (3) Disfluencies and repetition (e.g. speech rate, prolonged pauses); (4) Semantic coherence (e.g. content density, pronoun-to-noun ratio).

\subsubsection{Multi-Stage LLM reasoning pipeline for clinical interpretation}
To transform SHAP outputs and linguistic features into clinically meaningful interpretations, we developed a novel explanation generation system based on LLM orchestration. Inspired by multi-agent AI architectures \cite{azure_ai_agent_design_patterns}, our approach implemented a manually orchestrated sequential pipeline where specialized reasoning stages processed information in a predefined, linear order to achieve progressive interpretive refinement.

Our pipeline chained four specialized LLMs, each configured with distinct prompts and knowledge sources. All LLMs shared common information (the original transcript and model prediction) in their prompt. We employ LLaMA-3.1-70B-Instruct \cite{grattafiori2024llama} as the reasoning LLM across all four stages. The complete prompts for all four stages are available in the project \href{https://github.com/SpeechCARE/Cognitive-Speech-Explainer-Interspeech2026}{GitHub} repository.

\textbf{Stage 1 - token-level SHAP interpretation:} This LLM received SHAP scores as structured input (token, SHAP value) alongside the transcript and the model’s prediction (control, MCI, AD). An instruction prompt guided the LLM to map token importance to six cognitive-linguistic dimensions: lexical richness, syntactic complexity, disfluencies and repetition, semantic coherence, spatial reasoning difficulty, and executive function. To preserve interpretive fidelity, the prompt included clinical descriptions of each category. Providing these guidelines constrained the reasoning process within a cognitive-linguistic framework, reducing semantic drift and maintaining consistency with clinically grounded constructs.

\textbf{Stage 2 - feature-level linguistic interpretation:} Theory-informed linguistic features corresponding to the six cognitive-linguistic dimensions were provided as structured inputs, including their numeric values and operational definitions (e.g., Type–Token Ratio, mean length of utterance, pause frequency). When available, reference ranges from prior literature were included (e.g., Type–Token Ratio: 0–1, where lower values reflect lexical repetition). Providing these structured clinical descriptions ensured that the model’s reasoning remained grounded within the cognitive-linguistic domain and aligned with established clinical interpretations. The LLM was prompted to interpret these quantitative patterns in relation to cognitive-linguistic functioning. 

\textbf{Stage 3 - cross-source interpretation aggregation:} This stage received the outputs of Stages 1 and 2 as structured input and was prompted to compare SHAP-based token attributions with feature-level metrics, integrating convergent evidence into a single interpretation.

\textbf{Stage 4 - structured report generation:} The final stage in the sequence performed a "polish and summarize" workflow. The aggregated interpretation from Stage 3 was re-submitted with a summarization prompt instructing to extract the four most diagnostic linguistic categories from the six dimensions mentioned in stage 1. For each dimension, the model generated concise bullet-point explanations, providing a clinically interpretable report accessible to support clinical readability.

\subsection{Evaluation of clinical validity}
To assess the clinical validity of the explainability framework, two primary care physicians independently reviewed the final interpretations for a stratified subset of 70 English-language samples (30\% of cases correctly predicted by our screening model), balanced by diagnostic label, age, sex, and education. The evaluation was conducted under blinded conditions: during the independent assessment phase, physicians were not provided with the model’s predicted label, ground-truth diagnosis, or each other’s ratings.

The evaluation occurred in two phases: (1) iterative refinement of Stage 1 and Stage 2 prompts using a separate validation subset to ensure clinically meaningful categorization; and (2) independent evaluation of the finalized reports on the test data.

\subsection{Clinical utility and workflow integration study}
To evaluate practical applicability, we conducted a usability study with three primary care physicians and two neurologists who were not involved in the model development or prior validation phases. The assessment examined: (1) feasibility of integration into clinical workflows; and (2) perceived impact on diagnostic decision-making. Participants completed the System Usability Scale (SUS) and participated in structured \textit{think-aloud} sessions while using the interface. The think-aloud protocol captured real-time clinical reasoning, while SUS scores quantified the system's complexity, usability, and implementation readiness.
\begin{figure*}[!ht]
    \centering
    \begin{subfigure}{0.49\linewidth}
        \centering
        \includegraphics[width=\linewidth]{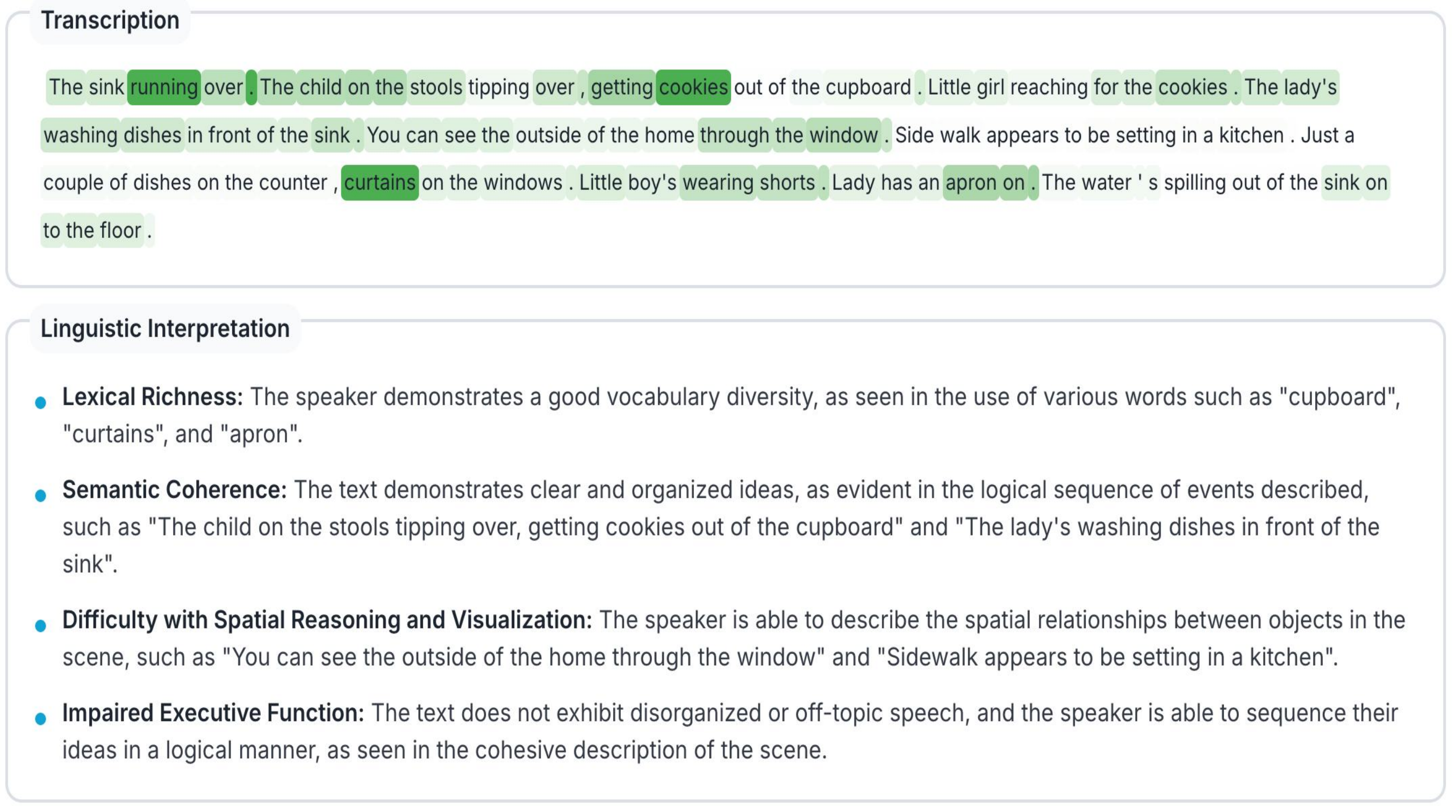}
        \caption{Control Case}
        \label{fig:Control}
    \end{subfigure}
    \hfill
    \begin{subfigure}{0.49\linewidth}
        \centering
        \includegraphics[width=\linewidth]{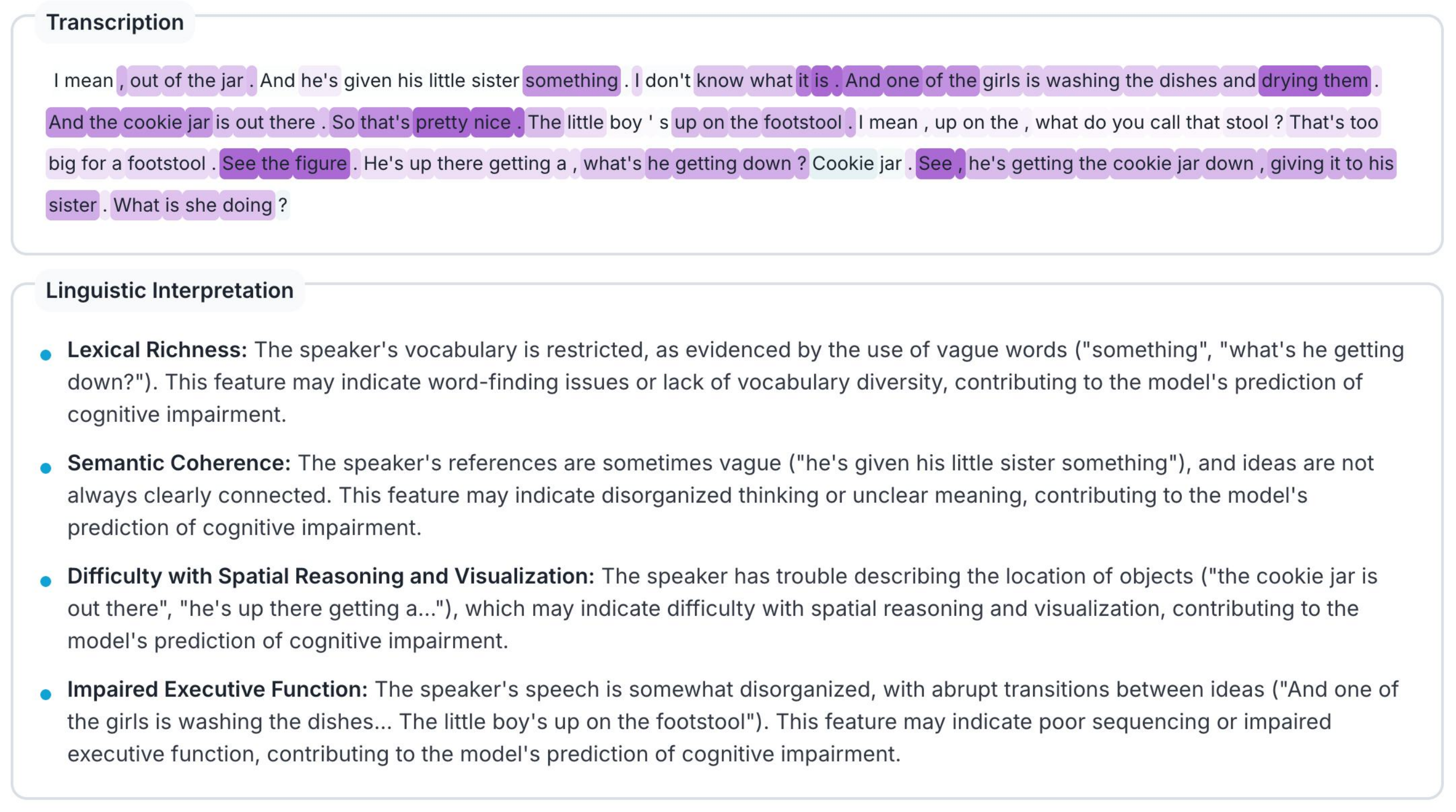}
        \caption{AD Case}
        \label{fig:Adrd}
    \end{subfigure}
    
    \caption{Explainability output comparison between (a) a Control and (b) an AD case. Each panel displays a word-level SHAP attribution map (top), where color intensity reflects token contribution to the model's prediction, and a structured linguistic interpretation report (bottom) generated by the LLM pipeline.}
    \label{fig:Result}
\end{figure*}

\section{Result}

\subsection{Model performance evaluation}
We began our evaluation by analyzing model performance. Following training, we selected the model checkpoint that achieved the highest F1-score on the validation set and used this checkpoint to evaluate performance on the official Test set released by the PREPARE challenge organizers. The model achieved an Area Under the Curve (AUC) of 86.83\% and an F1-score of 72.11\%. These findings confirm the model's balanced effectiveness in accurately detecting cognitive impairment.

\subsection{Explainability framework: SHAP and LLM-based interpretation}

Figure \ref{fig:Result} illustrates the explainability framework for representative control and AD cases. Each visualization consists of two aligned layers: (i) a word-level SHAP attribution map, where highlighting intensity reflects stronger token contribution to the model’s prediction; and (ii) a structured clinical report that translates these attributions into plain-language interpretations.

In Figure \ref{fig:Control} (control case), the SHAP map highlights a diverse vocabulary and complex sentence structure. The framework converted specific tokens, such as “curtains” and “apron”, into evidence for high lexical richness. Furthermore, the model mapped utterances such as "You can see the outside of the home through the window" to a preserved capacity for spatial reasoning and visualization, with the LLM correctly interpreting these as indicators of intact visuospatial description ability. 

In Figure \ref{fig:Adrd}, AD case, SHAP highlighted lexical access difficulty and referential ambiguity, including vague expressions such as “something” and “what’s he getting down?”. The report prioritized impairments in lexical richness, semantic coherence, executive function, and spatial reasoning, showing translation of model attributions into clinically meaningful explanation.

\subsection{Clinical validation of interpretations}
Two primary care physicians independently evaluated the structured reports under blinded conditions, without access to model predictions or each other’s assessments. The structured reports were found to align closely with each patient's individual cognitive-linguistic biomarkers, with the LLM successfully mapping SHAP-highlighted tokens to the specific impairment profile of the speaker rather than producing generic or population-level descriptions. Quantitatively, physicians agreed on 98\% of cases, with Cohen’s $\varkappa$ = 0.85.

The high agreement rate between physician assessments and LLM-generated reports underscores the potential of sequential LLM orchestration as a viable mechanism for bridging the gap between black-box transformer predictions and individualized, biomarker-grounded clinical narratives.

\subsection{Usability and clinical workflow integration}

Clinicians participated in "think-aloud" sessions, verifying that the interface has strong potential for integration into clinical workflows. The clinicians reported that the provided linguistic summaries offer actionable evidence that can enhance decision-making for the patient evaluation for cognitive impairment. Analysis of the System Usability Scale (SUS) responses achieved an overall usability score of 82 out of 100, showing that the system is easy to use, well-integrated, and quickly learnable without the need for extensive technical support.

\section{Discussion}
This paper is the first to present a multi-stage explainability framework showing that sequentially combining SHAP attributions, theory-informed linguistic features, and LLM-based reasoning can bridge the gap between black-box transformer predictions and clinically meaningful narratives. Speech analysis provides a functional complement to recently approved biomarker technologies such as plasma assays (pTau217 and $\beta$-amyloid approved by FDA in 2025), which reflect neuropathology but do not capture its functional impact on communication. In contrast, speech quantifies everyday cognitive-linguistic function through prosody, acoustic parameters, and semantic organization. Clinician evaluation supports the feasibility and clinical utility of individualized explanations within diagnostic workflows, positioning sequential LLM orchestration as a promising paradigm for XAI-enabled cognitive screening.

Despite these promising results, several limitations warrant attention. The current framework focuses on linguistic explainability; future work should extend to acoustic transformer explanations by translating mHuBERT attention patterns and acoustic representations into clinically interpretable narratives (e.g., pitch, prosody, speech rate) to fully leverage the multimodal SpeechCARE-AGF model.

\section{AI disclosure}
We used ChatGPT (OpenAI, GPT-4) to assist proofreading the manuscript for language quality. The content generated by this tool was thoroughly reviewed, revised, and approved by the authors. No AI tools were used for drafting the manuscript, data analysis, experiment design, or results interpretation.

\bibliographystyle{IEEEtran}
\bibliography{mybib}

\end{document}